\documentclass{article}
\usepackage{arxiv}

\usepackage{mathtools}
\usepackage[ruled, linesnumbered]{algorithm2e}
\usepackage[noend]{algpseudocode}
\usepackage{graphicx}
\usepackage{hyperref}
\usepackage{subcaption}

\title{Autonomous sPOMDP Environment Modeling With Partial Model Exploitation}

\author{ \href{https://orcid.org/0000-0002-8206-2031}{\includegraphics[scale=0.06]{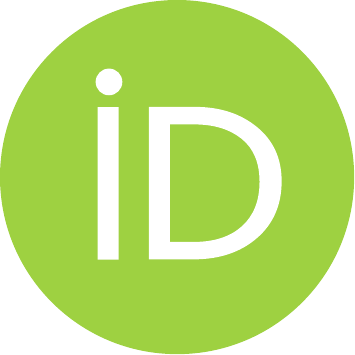}\hspace{1mm}Andrew~Wilhelm}\thanks{These two authors contributed equally} \\
	EpiSys~Science,~Inc.\\
	Poway, CA, 92064\\
	\texttt{ajw343@cornell.edu} \\
	\And
	\href{https://orcid.org/0000-0003-0217-2554}{\includegraphics[scale=0.06]{orcid.pdf}\hspace{1mm}Aaron~Wilhelm}\footnotemark[1] \\
	EpiSys~Science,~Inc.\\
	Poway, CA, 92064\\
	\texttt{ajw344@cornell.edu} \\
	\And
	\href{https://orcid.org/0000-0002-9419-5042}{\includegraphics[scale=0.06]{orcid.pdf}\hspace{1mm}Garrett~Fosdick} \\
	EpiSys~Science,~Inc.\\
	Poway, CA, 92064\\
	\texttt{gfosdick@episci.com} \\
}

\hypersetup{
pdftitle={Autonomous sPOMDP Environment Modeling With Partial Model Exploitation},
pdfsubject={q-bio.NC, q-bio.QM},
pdfauthor={David S.~Hippocampus, Elias D.~Striatum},
pdfkeywords={POMDP, Surprise-Based Learning, SLAM},
}

\begin{document}
\maketitle

\begin{abstract}
	A state space representation of an environment is a classic and yet powerful tool used by many autonomous robotic systems for efficient and often optimal solution planning. However, designing these representations with high performance is laborious and costly, necessitating an effective and versatile tool for autonomous generation of state spaces for autonomous robots. We present a novel state space exploration algorithm by extending the original surprise-based partially-observable Markov Decision Processes (sPOMDP), and demonstrate its effective long-term exploration planning performance in various environments. Through extensive simulation experiments, we show the proposed model significantly increases efficiency and scalability of the original sPOMDP learning techniques with a range of 31-63\% gain in training speed while improving robustness in environments with less deterministic transitions. Our results pave the way for extending sPOMDP solutions to a broader set of environments.
\end{abstract}



\section{\uppercase{Introduction}}
\label{sec:introduction}
	
\noindent As robotics systems become more autonomous, there is a rising demand for robots to be able to navigate in increasingly complex and uncertain real-world environments. Many of these environments are partially observable, in that the agent cannot perfectly observe all states of an environment, and stochastic, in that an agent’s actions and observations are probabilistic in nature. Thus, in developing algorithms for state estimation, localization, and mapping in robotics platforms, the partially observable and stochastic nature of these environments must be considered as attempting to predict an agent’s movement in the environment becomes a non-trivial problem.
    
One approach to represent these environments is to model the environment as a Partially Observable Markov Decision Process (POMDP)~\cite{kaelbling1998planning}. A POMDP model $\mathcal{M}$ attempts to model an underlying environment $\mathcal{E}$ using a discrete set of model states $M$, a discrete set of actions $A$, a discrete set of transition probabilities $T$, a discrete set of observations $O$, and a discrete set of observation probabilities $\Omega$. Using a POMDP model to predict an agent’s trajectory is desirable as an extensive set of POMDP solutions already exist to solve planning and decision-making problems, but to build an accurate POMDP representation of a stochastic environment with latent states (states that share the same most likely observation) is a challenging task. A second approach to develop robust algorithms in a partially observable, stochastic environment is predictive methods. Predictive methods \cite{Collins2017,Littman2017} can be advantageous as simple counting methods can be used to estimate the probabilities of events occurring. However, unlike POMDPs, predictive methods do not necessarily define a relationship between the underlying environmental structure and their predictive experiments.

In this paper, we propose a novel algorithm to significantly increase the speed and robustness of the generation of surprise-based partially-observable Markov Decision Processes (sPOMDP) models, a type of POMDP model that attempts to address the trade-offs between predictive and model-based approaches. sPOMDP models are advantageous in that computationally inexpensive counting algorithms can be used to identify latent states in the environment and ultimately build a POMDP model of the underlying environmental structure.  

Our key contribution is an enhanced and highly versatile sPOMDP generation methodology, proven with extensive simulation, that is capable of significantly increasing the training speed and scalability of pre-existing sPOMDP active environment learners while gaining greater model generation robustness simultaneously. We achieve this by building sPOMDP models with a novel exploration algorithm that plans routes through the environment to more quickly learn an accurate model. Additionally we provide a novel transition posterior updating method for faster and more accurate transition estimates. The proposed algorithm achieves 31-63\% gain in training speed without losing the state representation accuracy. Our results pave the way for extending sPOMDP solutions to a broader set of environments.


\section{\uppercase{Related Work}}
\label{sec:relatedwork}
\noindent Our paper addresses the longstanding issue of ``autonomous learning from the environment'' which was defined as early as 1993~\cite{Shen94autonomouslearning}. This involves an agent that must learn a discrete environment with limited partial observations that can cause latent states to be hidden by other states with identical observations. Additionally the agent is limited to its own actions to explore and gather data with no ability to reset to get to a new state.

Addressing latent states in off-model approaches has been addressed by providing history to reinforcement learning agents using Recurrent Neural Networks (RNN), and Long Short Term Memory (LSTM), and other similar approaches. This permits them to use previous states to differentiate states with identical observations. These are powerful approaches, but require a large amount of hyperparameter and architecture knowledge to create a model capable of learning. Collins provides a Stochastic Distinguishing Experiment based approach that uses a trajectory history to localize a state and differentiate identical states~\cite{Collins2019}. This approach is robust to hyperparameter sensitivity with non-parametric scaling. However, this work is also inefficient and scales poorly since it often repeats redundant transitions that gain no new state or transition knowledge.

The majority of our work goes into creating more intelligent exploration and planning for an agent with built-in curiosity.  There are two areas of research closely related to this goal. These areas are intrinsic motivation~\cite{aubret2019survey,10.3389/neuro.12.006.2007,5471106,NIPS2015_5668,4141061} and reinforcement learning via artificial curiosity~\cite{DBLP:journals/corr/abs-1905-09275,Storck95reinforcementdriven,10.3389/fnbot.2013.00025,Pathak_2017_CVPR_Workshops}. These approaches seek out novel situations to learn generally applicable skills.


\section{\uppercase{Background}}
\label{sec:background}

\subsection{Surprise Based Learning (SBL) of $\alpha \epsilon$-POMDP Environments}
\label{subsec:learningPOMDP}

\noindent An $\alpha \epsilon$-POMDP environment is a subset of POMDP environments in which the transition and observation probabilities have a specific distribution as follows~\cite{Collins2018}. For each starting model state $m$ under action $a$, the most likely transition occurs with probability $\alpha$, and the remaining probability distribution is equally distributed across all other ending model states. The observation probabilities follow a similar distribution where the most likely observation $o$ at a state $m$ is observed with probability $\epsilon$, and the remaining probability distribution is equally distributed across all other observations. Approximating a POMDP model to an $\alpha \epsilon$-POMDP environment is a useful approximation as it simplifies certain calculations and it has been demonstrated that algorithms designed for $\alpha \epsilon$-POMDP environments can generalize to non $\alpha \epsilon$-POMDP environments~\cite{Collins2019}.

To learn unknown $\alpha \epsilon$-POMDP environments with surprise-based learning, Collins used an iterative method where an sPOMDP model is generated, the sPOMDP transition probabilities are updated through a counting method as the agent explores the environment, and then a new sPOMDP model is generated if it is determined latent states exist in the model~\cite{Collins2018}. An sPOMDP is a hybrid latent-predictive model that uses the same parameters ($M$, $A$, $T$, $O$, $\Omega$) from a traditional POMDP and an additional parameter $S$, a discrete set of Stochastic Distinguishing Experiments (SDEs). Each SDE $s \in S$ is an ordered set of observations and actions in a stochastic, partially observable environment that can be used to distinguish latent states in the environment~\cite{Collins2017}. More formally, each SDE $s$ has a set of associated model states $\{m_s\}$ that, upon execution of $s$, the model state $m^i_s$ can be disambiguated by its outcome sequence $g_{m^i_s}$. SDEs are generated so that if an agent traversing the environment successfully makes observations and takes actions that perfectly correspond to an SDE, then the agent most likely started at a specific model state with a certain probability.

As the layout of the environment is initially unknown, the initial set of SDEs in the sPOMDP model is the set of possible observations. During each iteration, the agent determines if latent environment states exist with the current sPOMDP model. If latent states do exist, the agent determines from its experimentation which model state represents multiple environment states and how to split that model state into two new model states to better represent the latent environment states. Through this method, the agent is able to simultaneously learn both the model states and their transitions for an sPOMDP environment.
	
To learn the sPOMDP transition probabilities, the agent must generate a policy, a set of ordered actions for it to execute, that allows it to remain localized while still fully exploring the environment. The policy uses the set of actions from the SDEs that corresponds to model states ${m_s}$ that have the same initial observation as the agent’s current observation. With probability $explore$, the actions of the SDEs can be replaced by random actions instead. As the agent performs these actions from the policy they are removed from the policy. When the policy is empty, the process repeats and a new SDE or random actions are added to the policy. This continues until $numActions$ actions are taken, where $numActions$ is a user-defined hyperparameter specifying the maximum number of actions the agent should take.

The transition posteriors are updated using a counting method. The transition counts $\gamma^{mam'}$ are updated each iteration using Equation~\ref{eqn:collins_posterior}, where $\gamma^{mam'}_{t-1}$ is the Dirichlet hyperparameter from the previous iteration that represents a soft count of the number of times the agent has taken action $a$ from state $m$ and ended in state $m'$ and the function $\textbf{1}_{o\equiv m}(m')$ is an indicator function that is $1$ if and only if the first observation of $m'$ matches the agent’s current observation $o$. The update is proportional to the likelihood of the transition occurring and normalized across the other $\gamma^{ma}$ updates.

\begin{equation}\label{eqn:collins_posterior}
\gamma^{mam'}_{t} = \gamma^{mam'}_{t-1} + \eta \textbf{1}_{o\equiv m'}(m') T^{mam'}_{t-1}b_{\mathcal{M},t-1}(m)
\end{equation}

Next, the algorithm attempts to identify if latent states exist in its model and how to split the model accordingly. The associated gain values $G_{ma}$ of each of model state-action pair $(m, a)$ in $M\times A$ are calculated where the value of $G_{ma}$ is the reduction in entropy the model can achieve by incorporating one more time step in its history. A new SDE is generated by concatenating the first observation of $m$, the action $a$, and then the associated outcome sequences of $m_1$ and $m_2$, the two most likely model states to transition into if the agent were to be in model state $m$ and take action $a$. A new sPOMDP model is created with this newly generated SDE, and then the algorithm repeats and begins to learn the new sPOMDP transition probabilities.

Once the model is complete, the agent can then utilize the model to travel to a state with a goal observation, using an algorithm introduced by Collins~\cite{Collins2019}.


\section{\uppercase{Approach}}
\label{sec:approach}

\subsection{Optimized Exploration and Navigation with Agent Control}
\label{subsec:approach:agentControl}
	
	\noindent Previous approaches to modeling $\alpha \epsilon$-POMDP environments require a large number of actions to generate a model of the environment. This occurs for two reasons. First, no early termination conditions are defined; instead, the agent continues to perform SDEs and random actions until a predefined number of actions are performed. It is, however, non-trivial to determine a minimum number of total actions that still yields a model with high accuracy as this number greatly depends upon the model and underlying environment.
	
\begin{figure*}[!h]
	\centering
	\begin{subfigure}[b]{0.28\linewidth}
		\includegraphics[width=\linewidth]{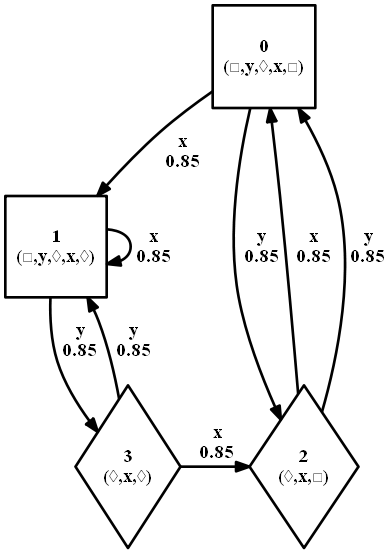}
		\caption{$\alpha \epsilon$-Shape Environment}
		\label{fig:s_env}
	\end{subfigure}
	\begin{subfigure}[b]{0.31\linewidth}
		\includegraphics[width=\linewidth]{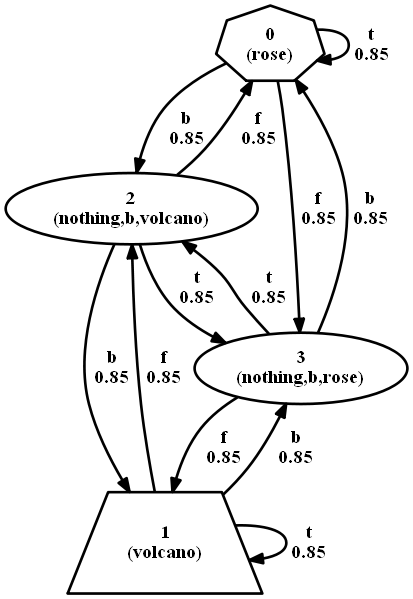}
		\caption{$\alpha \epsilon$-Little Prince Environment}
		\label{fig:lp_env}
	\end{subfigure}
	\begin{subfigure}[b]{0.38\linewidth}
		\includegraphics[width=\linewidth]{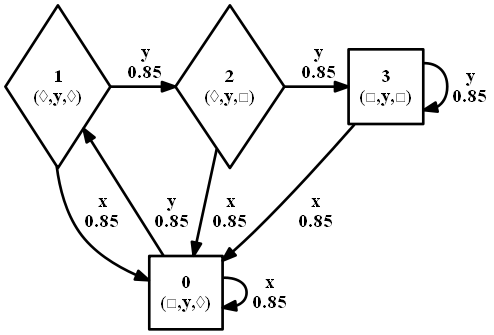}
		\caption{$\alpha \epsilon$-Balance Beam Environment}
		\label{fig:bb_env}
	\end{subfigure}
	\caption{Above are the testing environments. These environments are all POMDP environments, with the arrows indicating the most likely state transition under the corresponding action. The most likely observation in each state is represented by the shape of that state. Each state contains a valid SDE (below the state number) that could be used in a sPOMDP model to represent that state.}
	\label{fig:environments}
\end{figure*}

	The second reason that previous modeling approaches require a large number of actions is that they rely upon chance to get to all model states. For certain environments, such as the $\alpha \epsilon$-Balance Beam environment (Figure~\ref{fig:bb_env}), some states are difficult to reach by chance only. Thus, to ensure that transitions from these states are accurately determined, the agent must perform a large number of actions.
	
	To reduce the number of actions required to learn an $\alpha \epsilon$-POMDP environment, we present a navigation policy (Algorithm~\ref{alg:navPol}) that guides the agent to take actions that help it learn its environment quicker. Overall, the objective of the navigation policy is to have the agent learn the environment transitions as quickly as possible up to a given confidence level. We define the confidence of a transition in Equation~\ref{eqn:confidence} as the summation of the gammas of the Dirichlet distribution for that transition, divided by the number of model states (line~\ref{NP:ln:confidences}). Since the values of the Dirichlet gammas are all initialized to one, dividing by the number of model states helps normalize the confidence value across differently sized models.
	
\begin{equation}\label{eqn:confidence}
confidence(m,a) := \left(\sum\limits_{m' \in M} \gamma^{mam'}\right)\ / \ |M|
\end{equation}

\begin{algorithm*}[!t]
\DontPrintSemicolon
\KwIn{$\mathcal{M}$: the model, $confidence\_factor$: the confidence factor}
\KwOut{$policy$: the list of actions for the policy}

$policy := \{\}$\;
$nonzero\_values \gets \text{the \# of nonzero values in } \textit{$\mathcal{M}$.beliefState}$\;
\tcc{Check if we need to localize the agent}
\uIf {\upshape $agent.performedExperiment {\bf \ and} \text{\ entropy} \textit{($\mathcal{M}$.beliefState, base=nonzero\_values)} > localization\_threshold$\label{NP:ln:if_localization}}{
		$policy.append(\mathcal{M}.getSDEActions())$\;\label{NP:ln:SDE}
		\uIf{$agent.performedExperiment$\label{NP:ln:start_rand}} {
            \tcc{Perform random actions in case latent states exist}
			$randomActions := (size(\mathcal{M}.SDESet) - 1) \text{ random actions}$\;
			$policy.append(randomActions)$\;
			$agent.performedExperiment \gets False$\;\label{NP:ln:end_rand}
		}}
	\uElse{
			\tcc{Try to perform experiment}
			$confidences := \text{(sum of each row of } \mathcal{M}.totalCounts)\ / \ size(\mathcal{M}.SDESet)$\label{NP:ln:confidences}\;
			\For {$a \in \mathcal{M}.actionSet$}{
				\uIf {$confidences[a, agent.currentState] < confidence\_factor$\label{NP:ln:start_exp}}{
					$policy.append(a)$\;
					$agent.performedExperiment \gets True$\;
					\textbf{break}\label{NP:ln:end_exp}
				}
				\tcc{Try to get to a state to do an experiment}
				\uIf {$agent.performedExperiment == False$}{
					\tcc{Algorithm~\ref{alg:getPath}}
					$path := getPathToExperiment(\mathcal{M}, confidence\_factor)$\;\label{NP:ln:getPath}
					$policy.append(path[0])$\;
				}
			}
		}
\Return {$policy$}\;
\caption{Navigation Policy\label{alg:navPol}}
\end{algorithm*}
	
The navigation policy consists of three possible stages. If the agent is unsure of its localization, it will localize by performing an SDE (line~\ref{NP:ln:SDE}). The agent’s certainty in its belief state is quantified by calculating the entropy of the agent’s belief state (line~\ref{NP:ln:if_localization}). If the agent is certain of its localization, it will then move to the experimentation stage.

During the experimentation stage the agent tries to perform actions from states that have transitions with low confidence values. If the agent is at such a state, it will perform that action experiment and set $agent.performedExperiment$ to $True$  (lines~\ref{NP:ln:start_exp}-\ref{NP:ln:end_exp}). Then, the next time the policy is updated, the agent relocalizes itself to maximize the update to the transition gammas, and then the agent performs $|M - 1|$ random actions (lines~\ref{NP:ln:start_rand}-\ref{NP:ln:end_rand}). In the case that latent states exist, these random actions help make sure that the agent navigates through hidden states. If the agent has knowledge that no latent states exist, then these random actions can be disregarded.

\begin{figure*}[!tbhp]
  \centering
  \includegraphics[width=0.3\linewidth]{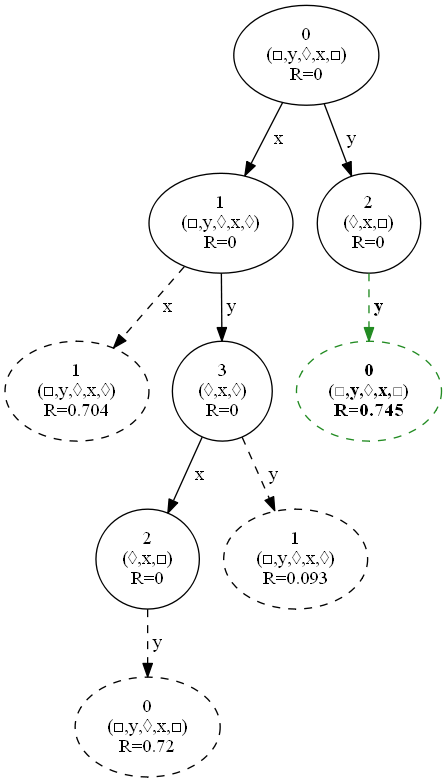}
  \caption{The tree used in the navigation policy. Each node represents a different sPOMDP model state (in this case numbered 0-3), and contains the SDE for that state as well as that node's calculated reward. The SDEs are composed of a sequence of observations (diamonds and squares) and actions (x and y). The dashed edges are experiments and dashed nodes are the expected resulting state after the experiment is performed. The green node and edge is the experiment with the highest reward.}
  \label{fig:navTree}
\end{figure*}

If the agent is not in a state where an experiment can be performed, then the agent attempts to travel to one. Algorithm~\ref{alg:getPath} determines the actions, if any, that will lead the agent to a state that needs experimentation. To do so, a tree is constructed with the root node representing the current state, as shown in Figure~\ref{fig:navTree}. Each node has a reward value; if an experiment was performed in the immediate step before, the reward value is calculated using Equation \ref{eqn:reward} (line~\ref{GP:ln:reward}), otherwise the reward value is zero. Calculating the reward as shown in Equation~\ref{eqn:reward} encourages the agent to take more confident transitions to states where experiments need to be done, and to encourage the agent to finish learning transitions that are close to surpassing the $confidence\_factor$ so it can use those confident transitions in the future. Additionally, each node has a probability value that is the probability of accurately getting to that state with the given actions.

\begin{equation}\label{eqn:reward}
\begin{gathered}
\begin{split}
reward(node) =\quad node.parent.probability \times \frac{confidence(m,a)} {confidence\_factor - 1}
\end{split}
\end{gathered}
\end{equation}

The tree is constructed layer by layer in the following manner. For each of the leaf nodes in the previous layer, the most likely resulting state from each action is added as a node if 1.) there have been no experiments on this path yet, 2.) this state has not been added to the path yet, and 3.) if the number of experiments for this transition exceeds the confidence factor, then entropy of the nonzero values of the belief state must be less than the $localization\_factor$ (lines~\ref{GP:ln:each_node}-\ref{GP:ln:end_each_node}). Condition 3 ensures that for non-experimental transitions, the agent can enter the destination state with a high degree of localization, which may not be possible if latent states exist. Construction continues in this manner until a depth of $|M|$ is reached, as at this depth the agent should be able to reach any other model state through the tree. Once the tree is constructed, Algorithm~\ref{alg:getPath} returns the actions that correspond to the path to the highest reward node, if a nonzero reward node exists.

Once the path to the node with the highest reward is determined, the agent adds the first action of the path to its policy. Only the first action is added because it is possible that the agent attempts to travel to a state, but instead with error $1-\alpha$ travels to another, rendering the rest of the actions in the path inaccurate. With this in mind, for subsequent calls for the policy, the path to the highest reward node is recalculated to ensure that the agent is always making optimal decisions.

The navigation policy ends once one of two termination conditions is met. In the ideal case, experimentation ends when the number of experiments for each $(m, a)$ pair has exceeded the confidence factor. However, this is not always possible when latent states exist. When latent states exist, the transition probabilities of the model may result in states needing experimentation that cannot be reached reliably, i.e. states that cannot be reached at all, or states that can only be reached via transition(s) that distribute the belief state too much. In this case, the agent reverts back to the traditional policy of random actions and SDEs until one of the model states can be split upon.

Besides a difference in objective, our navigation policy for training differs than Collins'~\cite{Collins2019} algorithm to guide the agent to a goal observation in several ways. First, our algorithm is used during model learning and thus has to account for the uncertainty of transitions that are still being learned, unlike Collins'~\cite{Collins2019} algorithm which is used in the simpler case where model learning has already ended. Additionally, our tree-search approach incorporates the probability of the transitions and does a more comprehensive search than Collins'~\cite{Collins2019} greedy algorithm. Finally, it should be noted that in addition to the navigation algorithm, our approach also uses the modified transition posterior update equation that is discussed in Section~\ref{subsec:approach:posteriors}.

\begin{algorithm*}[!t]
\DontPrintSemicolon
\KwIn{$\mathcal{M}$: the model, $confidence\_factor$: the confidence factor}
\KwOut{$actions$: the actions to get to the experiment state with the most reward, null if no experimental states can be reached}
$transitionProbs := calculateTransitionProbs(\mathcal{M}.totalCounts)$\;
$confidences := sumEachRow(\mathcal{M}.totalCounts)\ / \ size(\mathcal{M}.SDESet)$\;
$statesOfInterest := \text{list of model states }m \text{ where $\exists$ an action }a \ni confidences[m,a] < confidence\_factor$\;

\uIf{$size(statesOfInterest == 0)$}{
	\Return $null$\;
}

$maxDepth := size(\mathcal{M}.SDESet)$\;
$root := new\ Node(state=agent.currentState, reward=0, prob=1, actions=\{\})$\;
$depth := 1$\;
$bestNode := null$\;

\tcc{construct tree level by level}
\While{$depth \leq maxDepth$} {
	\For{\upshape $node \in \text{ previous level}$}{
		\tcc{this means we already performed the experiment}
		\uIf{$node.reward \neq 0$}{\label{GP:ln:each_node}
			\textbf{continue}
		}
		\For{$a \in \mathcal{M}.actionSet$}{
			$row := transitionProbs[a,m,:]$\;
			$newActions := copy(node.actions).append(a)$\;
			$newProb := max(row)*node.prob$\;
			$destinationState := maxIndex(row)$\;
			\uIf{$confidences[a,node.state] \geq confidence\_factor$} {
				\uIf{\upshape $\text{\ entropy} \textit{($\mathcal{M}$.beliefState, base=nonzero\_values)} < localization\_factor {\bf \ and\ } destinationState \text{\ not in ancestor nodes}$} {
					$insertNode(reward=0, prob=newProb, parent=node)$\;
				}
			}
			\uElse{
				$newReward := node.prob*(confidences[a,node.state] / (confidence\_factor - 1))$\;\label{GP:ln:reward}
				$insertNode(reward=newReward, prob=newProb, parent=node)$\;
				\uIf{$bestNode==null {\bf \ or\ } reward > bestNode.reward$} {
					$bestNode \gets insertedNode$\;
				}
			}
		}\label{GP:ln:end_each_node}
	}
	
	\uIf{\upshape $\text{no nodes added this iteration}$} {
		\textbf{break}
	}
	$depth \gets depth + 1$\;
}

\uIf{$bestNode == null$}{
	\Return $null$
}
\uElse{
	\Return $\text{actions to get to bestNode}$
}
\caption{getPathToExperiment}
\label{alg:getPath}
\end{algorithm*}
\vspace*{-0.3cm}

\subsection{Transition Posterior Updates}
\label{subsec:approach:posteriors}
\noindent Under Equation~\ref{eqn:collins_posterior}, the transition updates are frequency dependent. In environments where certain latent states are visited much less frequently by the agent during learning, this frequency dependence can result in learning incorrect transition posteriors. Since an agent can only localize itself after successfully completing an SDE, then in cases where an agent does not successfully perform an SDE, its belief state can be distributed across multiple model states that have the same observation as the agent’s current observation. When updating the transition posteriors, all of these model states will have their corresponding transition gammas updated as well. This adversely affects the algorithm’s ability to learn the transition probabilities of transitions that are traversed less frequently as these transition probabilities may be incorrectly updated by other more frequent transitions that have the same starting observation. Thus, to more accurately update the transition posteriors for states that are visited less frequently, it is important to account for the agent’s relative confidence of being in a certain state prior to updating its transition posteriors.

To this end, we applied Equation~\ref{eqn:updated_posterior} to update the transition counts $\gamma^{mam'}$. This equation is nearly identical to the original posterior update equation (Equation~\ref{eqn:collins_posterior}) except that Equation~\ref{eqn:updated_posterior} only updates the transition posteriors for the most likely starting state. This prevents the agent from incorrectly updating the counts of less frequent transitions when it is unlikely the agent is actually in those states.

\begin{equation}\label{eqn:updated_posterior}
\begin{split}
\gamma^{mam'}_t = \gamma^{mam'}_{t-1} + (\eta \textbf{1}_{\text{argmax}_{m''}(b_{\mathcal{M},t-1}(m''))\equiv m}(m) \times\textbf{1}_{o\equiv m'}(m')  T^{mam'}_{t-1}b_{\mathcal{M},t-1}(m))
\end{split}
\end{equation}


 To achieve this, the function $\textbf{1}_{\text{argmax}_{m''}(b_{\mathcal{M},t-1}(m''))\equiv m}(m)$ is used in Equation~\ref{eqn:updated_posterior}. This function is an indicator function that is 1 if and only if the maximum value of the agent’s belief state occurs at model state $m$ and nowhere else. This ensure that if the maximum belief state value occurs at two or more states, then no transition posteriors are updated as the algorithm risks learning the incorrect transition.



\label{subsec:environments}

\section{\uppercase{Experiments and Results}}
\label{sec:experimental}

\subsection{Testing Environments}
The experiments below were conducted on a set of testing environments. The $\alpha \epsilon$-Shape environment and the $\alpha \epsilon$-Little Prince environment from Collins \cite{Collins2018} and a novel $\alpha \epsilon$-Balance Beam environment were chosen for testing as these environments have varying degrees of hiddenness (i.e. how many latent states there are and how many actions can be used to distinguish these latent states) and provide a good indicator of how each algorithm performs in a variety of environments. These testing environments are illustrated in Figure~\ref{fig:environments}.

\subsection{Navigation Policy Results}
\label{subsec:withControl}

\begin{figure*}[t]
	\centering
	\begin{subfigure}[b]{0.38\linewidth}
		\includegraphics[width=\linewidth]{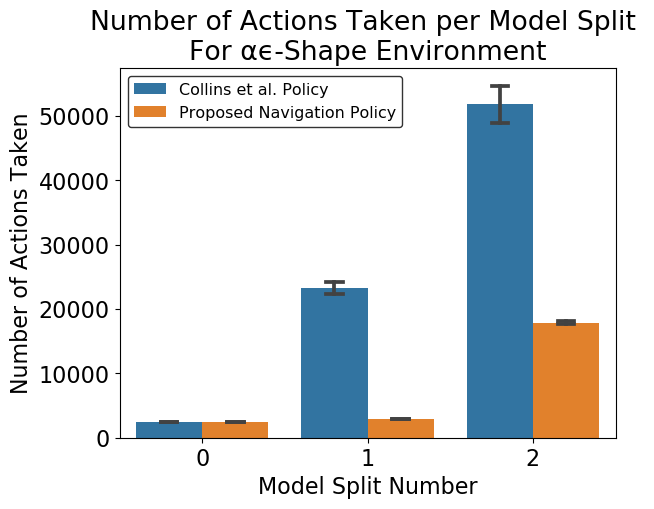}
		\caption{}
		\vspace*{.15cm}
		\label{fig:test2env2}
	\end{subfigure}
	\begin{subfigure}[b]{0.38\linewidth}
		\includegraphics[width=\linewidth]{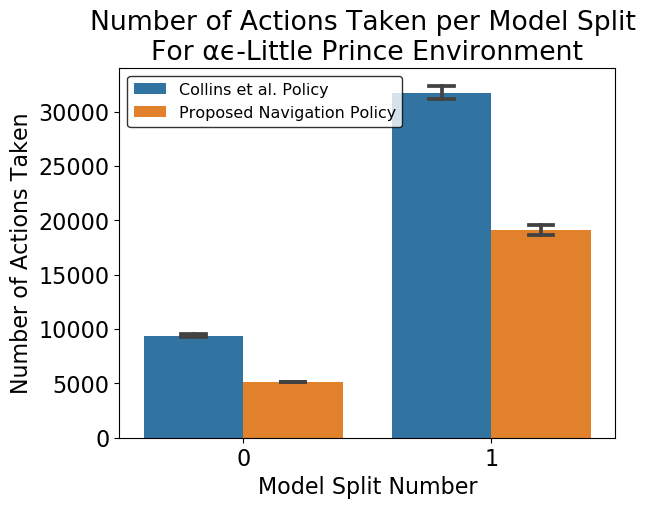}
		\caption{}
		\vspace*{.15cm}
		\label{fig:test2env3}
	\end{subfigure}
	\begin{subfigure}[b]{0.38\linewidth}
		\includegraphics[width=\linewidth]{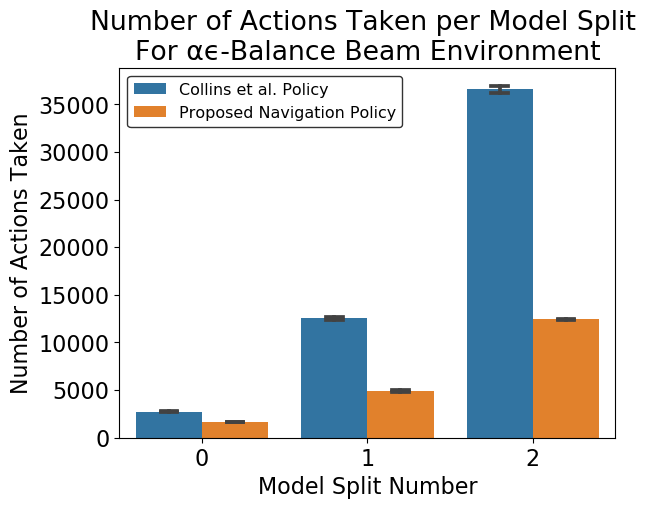}
		\caption{}
		\vspace*{.15cm}
		\label{fig:test2env7}
	\end{subfigure}
	\begin{subfigure}[b]{0.37\linewidth}
		\includegraphics[width=\linewidth]{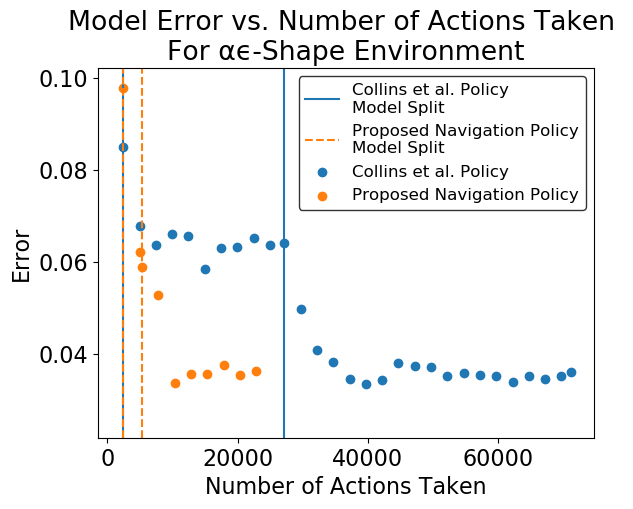}
		\caption{}
		\vspace*{.15cm}
		\label{fig:test2actions}
	\end{subfigure}
	\caption{The number of actions required to learn the environment for Collins et al. navigation policy and the proposed navigation policy for the $\alpha \epsilon$-Shape Environment (\ref{fig:test2env2}), the $\alpha \epsilon$-Little Prince Environment (\ref{fig:test2env3}), and the $\alpha \epsilon$-Balance Beam Environment (\ref{fig:test2env7}). The error bars represent the standard deviation of the ten trials. Overall, the proposed navigation policy is able to learn the environments in 31-63\% fewer actions than previous approaches. Figure~\ref{fig:test2actions} compares the error vs. number of actions taken between the two algorithms for the $\alpha \epsilon$-Shape Environment for a single trial.}
	\label{fig:test2}
\end{figure*}

\noindent Our first set of experiments are designed to demonstrate the improvements to the learning rate that our method of agent control can accomplish. Two sets of tests were conducted on various environments: one set of tests using randomly selected actions and SDEs (per Collins method of trajectory generation)~\cite{Collins2018} and a second set of tests using our proposed method of agent control, as described in Section~\ref{subsec:approach:agentControl}. Note that these sets of tests are identical other than the method of trajectory generation.

For each of these tests, $\alpha$ was set to 0.85 to ensure non-deterministic transitions and $\epsilon$ was set to 0.99 to preserve the $\alpha \epsilon$-environment properties. For Collins' method, the value of $explore$ was set to 0.5 as this allows the agent to remain localized while still exploring the environment. For our method, the localization threshold is set to 0.75 as that was found experimentally to be a good balance between keeping the agent localized while still enabling the agent to perform experiments. In both sets of tests early stopping (discussed in Section~\ref{subsec:approach:agentControl}) is enabled with the confidence factor set to 250, and the maximum number of actions to perform was set to 75,000. Additionally, both sets of tests had a gain threshold of 0.01 and a patience of 0. For each model split, the number of actions performed is recorded and averaged over 10 trials. Error is defined as the average of the absolute difference between the theoretical transitions and the sPOMDP model transitions.

Figure~\ref{fig:test2} shows the speed gains when the navigation policy is used as opposed to the random actions and SDEs approach. For all three environments tested ($\alpha \epsilon$-Shape Environment, the $\alpha \epsilon$-Little Prince Environment, and the $\alpha \epsilon$-Balance Beam Environment) the navigation policy uses 31-63\% fewer actions to learn the environment to the given confidence level. Figure~\ref{fig:test2actions} highlights these improvements for a single trial. For this trial, the proposed navigation policy was able to learn the environment with a lower error in less than half the number of actions that the policy proposed by Collins et al. took.

Our second set of experiments are designed to show the response of each of the algorithms to varying values of $\alpha$. Using the same testing parameters in the previous section, we performed 10 trials each for $\alpha$ values ranging from 0.65 to 0.99. Tests were performed in the $\alpha \epsilon$-Shape Environment.

\begin{figure*}[h]
	\centering
		\begin{subfigure}[b]{0.38\linewidth}
		\includegraphics[width=\linewidth]{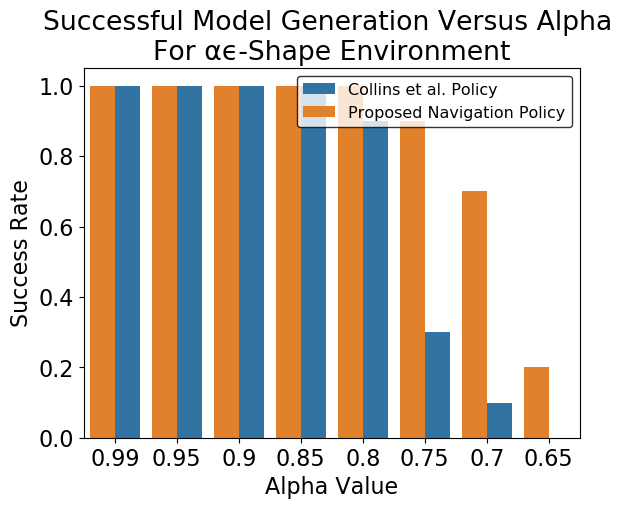}
		\caption{}
		\vspace*{.15cm}
		\label{fig:varyingAlphaAccuracy}
	\end{subfigure}
	\begin{subfigure}[b]{0.41\linewidth}
		\includegraphics[width=\linewidth]{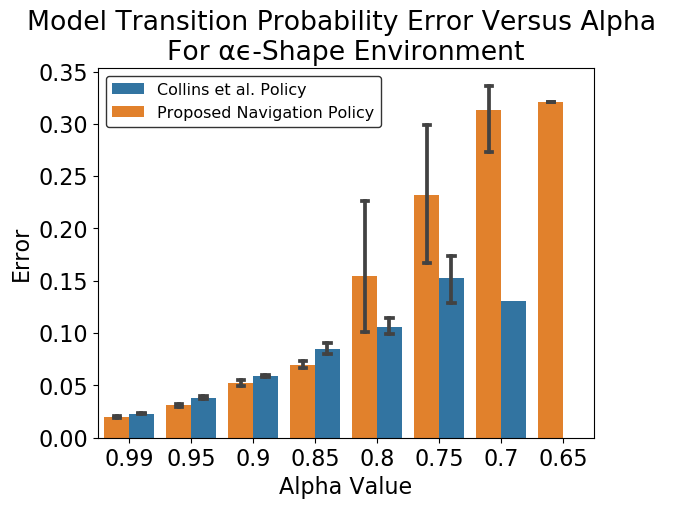}
		\caption{}
		\vspace*{.15cm}
		\label{fig:varyingAlphaError}
	\end{subfigure}
	\begin{subfigure}[b]{0.41\linewidth}
		\includegraphics[width=\linewidth]{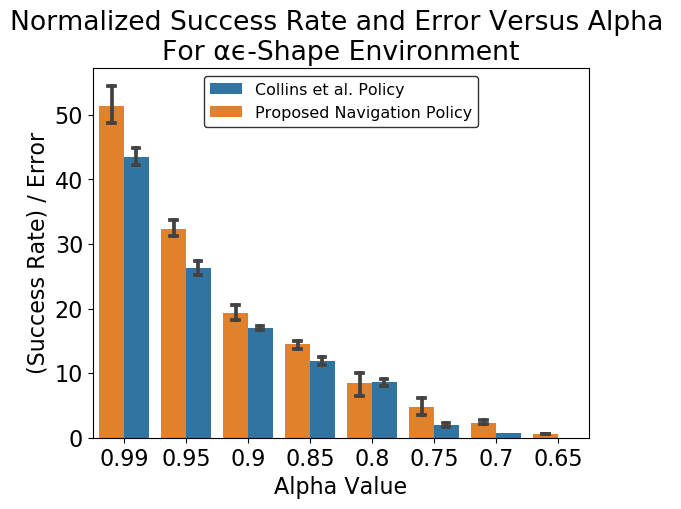}
		\caption{}
		\vspace*{.15cm}
		\label{fig:varyingAlphaRatio}
	\end{subfigure}
	\caption{A comparison of the Collins et al. navigation policy and the proposed navigation policy for the $\alpha \epsilon$-Shape Environment over varying values of $\alpha$. Results were collected over ten trials. Figure~\ref{fig:varyingAlphaAccuracy} compares the success rate of each policy in generating a POMDP model that perfectly corresponds to the states of the underlying POMDP model. Figure~\ref{fig:varyingAlphaError} compares the average error of the constructed models over varying values of $\alpha$. If a policy is unable to generate a correct model, then the error is not graphed.  Figure~\ref{fig:varyingAlphaRatio} compares the ratio between the model generation success rate and model error of the constructed models over varying values of $\alpha$. For both Figure~\ref{fig:varyingAlphaError} and Figure~\ref{fig:varyingAlphaRatio}, the error bars represent the standard deviation of the ten trials. Additionally, only trials that generated a model  perfectly corresponding to the underlying model are included in the data for these figures.}
	\label{fig:varyingAlpha}
\end{figure*}

Figure~\ref{fig:varyingAlpha} demonstrates our proposed navigation policy's improved robustness for sPOMDP model generation. For every value of $\alpha$ tested, our policy produced the same or a higher number of correct models. Additionally, we produce models that have lower levels of error at higher $\alpha$ values than Collins' method. Although Figure~\ref{fig:varyingAlphaError} may seem to suggest that the method proposed by Collins et al. has lower error at lower $\alpha$ values, as Figure~\ref{fig:varyingAlphaAccuracy} shows, for these values of $\alpha$ Collins' method often fails to generate a correct model at all. This is quantified in Figure~\ref{fig:varyingAlphaRatio}, where the ratio between the rate of successful model generations and the generated model error is used a normalization metric. Thus, in practice, our proposed method is more reliable.

The above tests can be replicated using the code repository at \href{https://github.com/EpiSci/SBL}{https://github.com/EpiSci/SBL}.


\section{\uppercase{Conclusion}}
\label{sec:conclusion}

\noindent We have presented a novel sPOMDP algorithm that allows an autonomous agent to quickly build a model of an $\alpha \epsilon$-POMDP environment if the agent’s actions are controllable. Additionally, we have empirically validated the improved robustness of our proposed navigation policy in environments with less deterministic transitions. We demonstrated the benefit of using these improvements to map $\alpha \epsilon$-POMDP environments over current methods via extensive simulation experiments. In the future, we plan on investigating the theoretical and computational feasibility of building models that use multiple SDEs to correspond to a single model state. This would allow the agent to localize more often and thus learn the transition probabilities quicker. Additionally, we intend to expand our testing to larger, non $\alpha \epsilon$-POMDP environments and develop methods that better generalize using sPOMDP models in non $\alpha \epsilon$-POMDP environments.


\bibliographystyle{plain}
\bibliography{bibliography}  

\end{document}